\pdfoutput=1
\documentclass[letterpaper]{article}
\usepackage{aaai18}
\usepackage{times}
\usepackage{helvet}
\usepackage{courier}
\usepackage{color}
\usepackage{amsmath}
\usepackage{amssymb} 
\usepackage{todonotes}

\newcommand{\bftheta}{{\boldsymbol \theta}}
\newcommand{\bfY}{{\bf Y}}
\newcommand{\bfK}{{\bf K}}
\newcommand{\bfZ}{{\bf Z}}

\title{Reversible Architectures for Arbitrarily Deep Residual Neural Networks}
\author{Bo Chang$^{*{1,2}}$, Lili Meng$^{*{1,2}}$, Eldad Haber$^{1,2}$, \\
\bf \Large{Lars Ruthotto$^{2,3}$, David Begert$^{2}$ and Elliot Holtham$^{2}$}\\
$^{1}$University of British Columbia, Vancouver, Canada. (bchang@stat.ubc.ca, menglili@cs.ubc.ca, haber@math.ubc.ca)\\
$^{2}$Xtract Technologies Inc., Vancouver, Canada. (david@xtract.ai, elliot@xtract.ai) \\
$^{3}$Emory University, Atlanta, USA. (lruthotto@emory.edu). This author is supported in part by NSF DMS 1522599.\\
$^{*}$Authors contributed equally.
}

\begin{document}
\maketitle
\begin{abstract}
Recently, deep residual networks have been successfully applied in many computer vision and natural language processing tasks, pushing the state-of-the-art performance with deeper and wider architectures. In this work, we interpret deep residual networks as ordinary differential equations (ODEs), which have long been studied in mathematics and physics with rich theoretical and empirical success. From this interpretation, we develop a theoretical framework on stability and reversibility of deep neural networks, and derive three reversible neural network architectures that can go arbitrarily deep in theory. The reversibility property allows a memory-efficient implementation, which does not need to store the activations for most hidden layers. Together with the stability of our architectures, this enables training deeper networks using only modest computational resources. We provide both theoretical analyses and empirical results. Experimental results demonstrate the efficacy of our architectures against several strong baselines on CIFAR-10, CIFAR-100 and STL-10 with superior or on-par state-of-the-art performance. Furthermore, we show our architectures yield superior results when trained using fewer training data.
\end{abstract}

\section{Introduction}
Deep learning powers many research areas and impacts various aspects of society \cite{lecun2015deep} from computer vision \cite{he2016deep,huang2016densely}, natural language processing \cite{cho2014learning} to biology \cite{esteva2017dermatologist} and e-commerce. Recent progress in designing architectures for deep networks has further accelerated this trend \cite{simonyan2014very,he2016deep,huang2016densely}. Among the most successful architectures are deep residual network (ResNet) and its variants, which are widely used in many computer vision applications \cite{he2016deep,pohlen2016full} and natural language processing tasks \cite{oord2016wavenet,xiong2017microsoft,wu2016google}. However, there still are few theoretical analyses and guidelines for designing and training ResNet.

In contrast to the recent interest in deep residual networks, system of Ordinary Differential Equations (ODEs), special kinds of dynamical systems, have long been studied in mathematics and physics with rich theoretical and empirical success \cite{coddington1955theory,simmons2016differential,arnolʹd2012geometrical}. The connection between  nonlinear ODEs and deep ResNets has been established in the recent works of \cite{weinan2017proposal,haber2017stable,haber2017learning,lu2017beyond,long2017pde,chang2017multi}. The continuous interpretation of ResNets as dynamical systems allows  the adaption of existing theory and numerical techniques for ODEs to deep learning.  For example, the paper \cite{haber2017stable} introduces the concept of stable networks that can be arbitrarily long. However, only deep networks with simple single-layer convolution building blocks are proposed, and the architectures are not reversible (and thus the length of the network is limited by the amount of available memory), and only simple numerical examples are provided. Our work aims at overcoming these drawbacks and further investigates the efficacy and practicability of stable architectures derived from the dynamical systems perspective.

In this work, we connect deep ResNets and ODEs more closely and propose three stable and reversible architectures. We show that the three architectures are governed by stable and well-posed ODEs. In particular, our approach allows to train arbitrarily long networks using only minimal memory storage. 
We illustrate the intrinsic reversibility of these architectures with both theoretical analysis and empirical results. The  reversibility property easily leads to a memory-efficient implementation, which does not need to store the activations at most hidden layers. Together with the stability, this allows one to train almost arbitrarily deep architectures using modest computational resources. 

The remainder of our paper is organized as follows. We discuss related work in Sec.~\ref{sec:rel}.  In Sec.~\ref{sec:meth} we review the notion of reversibility and stability in ResNets, present three new architectures, and a regularization functional. In Sec.~\ref{sec:exp} we show the efficacy of our networks using three common classification benchmarks (CIFAR-10, CIFAR-100, STL-10). Our new architectures achieve comparable or even superior accuracy and, in particular, generalize better when a limited number of labeled training data is used. In Sec.~\ref{sec:conc} we conclude the paper.

\begin{figure}
	\begin{center}
			\includegraphics[width = 1\linewidth]
            {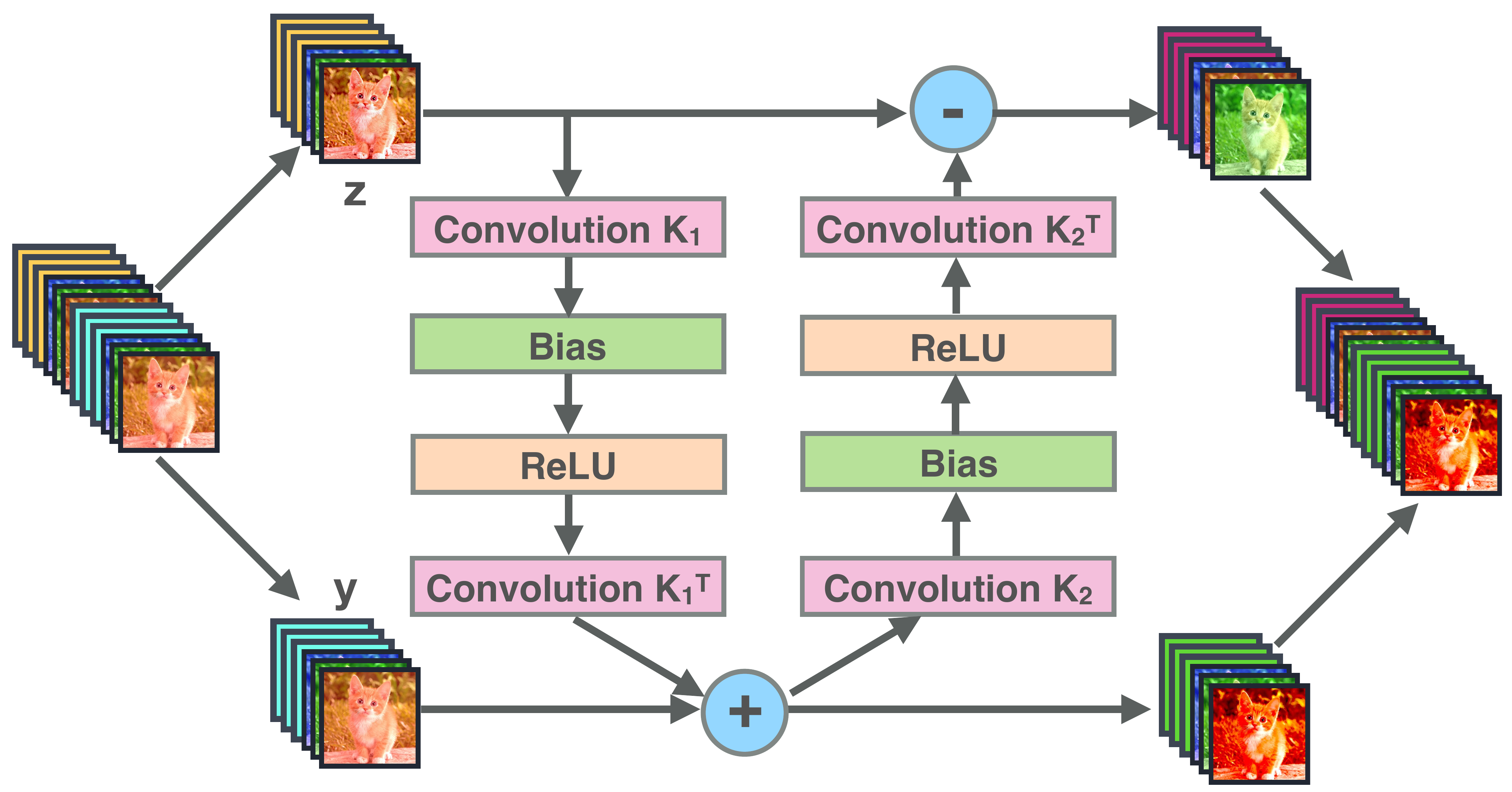}  
	\end{center}
	\caption{\textbf{The Hamiltonian Reversible Block.} First, the input feature map is equally channel-wise split to $\mathbf{Y}_j$ and $\mathbf{Z}_j$. Then the operations described in Eq.~\ref{eqn:hamiltonian-discretized} are performed, resulting in $\mathbf{Y}_{j+1}$ and $\mathbf{Z}_{j+1}$. Finally, $\mathbf{Y}_{j+1}$ and $\mathbf{Z}_{j+1}$ are concatenated as the output of the block.}
	\label{fig:hamiltonian-reversible-block}
\end{figure}

\section{Related Work}\label{sec:rel}

\subsection{Residual Neural Networks and Extensions}

ResNets are deep neural networks obtained by stacking simple residual blocks~\cite{he2016deep}. A simple residual network block can be written as
\begin{equation}
\label{equ:resnet-forward}
\mathbf{Y}_{j+1}=\mathbf{Y}_{j} + \mathcal{F}(\mathbf{Y}_j,\bftheta_j) \quad \mathrm{for} \quad j=0,...,N-1.
\end{equation}
Here, $\bfY_j$ are the values of the features at the $j$th layer and $\bftheta_j$ are the $j$th layer's network parameters. The goal of the training is to learn the network parameters $\bftheta$.
Eq.~\eqref{equ:resnet-forward} represents a discrete dynamical system.
An early review on neural networks as dynamical systems is presented in \cite{cessac2010view}.
%This interpretation allows one to use the rich theory and myriad algorithms that are available for dynamical systems arising from the discretization of ODEs to deep ResNets. 

ResNets have been broadly applied in many domains including computer vision tasks such as image recognition \cite{he2016deep}, object detection \cite{he2017mask}, semantic segmentation \cite{pohlen2016full} and visual reasoning \cite{perez2017learning}, natural language processing tasks such as speech synthesis \cite{oord2016wavenet}, speech recognition \cite{xiong2017microsoft} and machine translation \cite{wu2016google}. 

Besides broadening the application domain, some ResNet successors focus on improving accuracy \cite{xie2016aggregated,zagoruyko2016wide} and stability \cite{haber2017stable}, saving GPU memory \cite{gomez2017reversible}, and accelerating the training process \cite{huang2016deep}. For instance, ResNxt \cite{xie2016aggregated} introduces a homogeneous, multi-branch architecture to increase the accuracy. Stochastic depth \cite{huang2016deep} reduces the training time while increases accuracy by randomly dropping a subset of layers and bypassing them with identity function.

\subsection{Systems of Ordinary Differential Equations}

To see the connection between ResNet and ODE systems we add a hyperparameter $h>0$ to Eq.~\eqref{equ:resnet-forward} and rewrite the equation as
\begin{equation}
\label{equ:ODE1}
\frac{\mathbf{Y}_{j+1}-\mathbf{Y}_j}{h} = \mathcal{F}(\mathbf{Y}_j,\bftheta_j).
\end{equation}
For a sufficiently small $h$, Eq.~\eqref{equ:ODE1} is a forward Euler discretization of the initial value problem 
\begin{equation}
\label{equ:ODE}
\dot{\mathbf{Y}}(t)=\mathcal{F}(\mathbf{Y}(t),\bftheta(t)), \quad \mathbf{Y}(0) = \mathbf{Y}_0.
\end{equation}
Thus, the problem of learning the network parameters, $\bftheta$, is equivalent to solving a parameter estimation problem or optimal control problem involving the ODE system Eq.~\eqref{equ:ODE}. In some cases (e.g., in image classification), Eq.~\eqref{equ:ODE} can be interpreted as a system of Partial Differential Equations (PDEs).
Such problems have rich theoretical and computational framework, including techniques to guarantee stable networks by using appropriate functions $\mathcal{F}$, the discretization of the forward propagation process \cite{ap,ascherBook,bellman1953introduction}, theoretical frameworks for the optimization over the parameters $\bftheta$ \cite{bock1,SUlbrich_2002a,Gunz03}, and methods for computing the gradient of the solution with respect to $\bftheta$ \cite{bliss1919use}.

\subsection{Reversible Architectures}
Reversible numerical methods for dynamical systems allow the simulation of the dynamic going from the final time to the initial time, and vice versa. 
Reversible numerical methods are commonly used in the context of hyperbolic PDEs, where various methods have been proposed and compared \cite{nguyen2014five}. The theoretical framework for reversible methods is strongly
tied to issues of stability. In fact, as we show here, not every method that is algebraically reversible  is numerically stable. This has a strong implication for the practical applicability of reversible methods to deep neural networks.

Recently, various reversible neural networks have been proposed for different purposes and based on different architectures. Recent work by \cite{dosovitskiy2016inverting,mahendran2015understanding} inverts the feed-forward net and reproduces the input features from their values at the final  layers.  This suggests that some deep neural networks are reversible: the generative model is just the reverse of the feed-forward net \cite{arora2015deep}. \cite{gilbert2017towards} provide a theoretical connection between a model-based compressive sensing and CNNs. NICE \cite{dinh2014nice,dinh2016density} uses an invertible non-linear transformation to map the data distribution into a latent space where the resulting distribution factorizes, yielding good generative models. Besides the implications that reversibility has on the deep generative models, the property can be used for developing memory-efficient algorithms. For instance, RevNet~\cite{gomez2017reversible}, which is inspired by NICE, develops a variant of ResNet where each layer's activations can be reconstructed from next layer's. This allows one to avoid storing activations at all hidden layers, except at those layers with stride larger than one. We will show later that our physically-inspired network architectures also have the reversible property and we derive memory-efficient implementations.

\section{Methods}\label{sec:meth}
We introduce three new reversible architectures for deep neural networks
and discuss their stability. We capitalize on the link between ResNets and
ODEs to guarantee stability of the forward propagation process and the well-posedness of the learning problem. Finally, we present regularization functionals that favor smooth time dynamics.

\subsection{ResNet as an ODE}

Eq.~\eqref{equ:ODE} interprets ResNet as a discretization of a differential equation, whose parameters $\bftheta$ are learned in the training process.
The process of forward propagation can be viewed as simulating the nonlinear dynamics that take
the initial data, $\bfY_0$, which are hard to classify, and moves them to a final state
$\bfY_N$, which can be classified easily using, e.g., a linear classifier.

A fundamental question that needs to be addressed is, \emph{under what conditions is forward propagation well-posed?}
This question is important for two main reasons. First, instability of the forward propagation
means that the solution is highly sensitive to data perturbation (e.g., image noise or adversarial attacks). Given that
most computations are done in single precision, this may cause serious artifacts
and instabilities in the final results.
Second, training unstable networks may be very difficult in practice and, although
impossible to prove, instability can add many local minima.

Let us first review the issue of stability. A dynamical system is stable if
a small change in the input data leads to a small change in the final result.
To better characterize this, assume a small perturbation,
$\delta \bfY(0)$ to the initial data $\bfY(0)$ in Eq.~\eqref{equ:ODE}.
Assume that this change is propagated throughout the network. \emph{The question
is, what would be the change after some time $t$, that is, what is
$\delta \bfY(t)$?}

This change can be characterized by the Lyapunov exponent~\cite{lyapunov1992general}, which measures
the difference in the trajectories of a nonlinear dynamical system 
given the initial conditions. The Lyapunov exponent, $\lambda$, is defined as
the exponent that measures the difference:
\begin{equation}\label{eq:Lyapunov}
\|\delta \bfY(t)\| \approx \exp(\lambda t)\|\delta \bfY(0)\|.	
\end{equation} 
The forward propagation is well-posed when $\lambda \le 0$, and ill-posed if $\lambda>0$.
A bound on the value of $\lambda$ can be derived from the eigenvalues of the Jacobian matrix of $\mathcal{F}$ with respect to $\bfY$, which is given by
\begin{equation*}
\mathbf{J}(t) = \nabla_{\mathbf{Y}(t)} \mathcal{F}(\mathbf{Y}(t)).
\end{equation*}
A sufficient condition for stability is
\begin{equation}
\label{equ:RJacobian}
\max\limits_{i=1,2,...,n} Re(\lambda_i(\mathbf{J}(t))) \le 0, \quad \forall t\in[0, T],
\end{equation}
where $\lambda_i(\mathbf{J})$ is the $i$th eigenvalue of $\mathbf{J}$, and $Re(\cdot)$ denotes the real part.

This observation allows us to generate networks that are guaranteed to be stable. It should be emphasized that the stability of the forward propagation is necessary to obtain stable networks that generalize well, but not sufficient. In fact, if the  real parts of the eigenvalues in Eq.~\eqref{equ:RJacobian} are negative and large, $\lambda \ll 0$, Eq.~\eqref{eq:Lyapunov} shows that differences in the input features decay exponentially in time. This complicates the learning problem and therefore we consider architectures that lead to Jacobians with (approximately) purely imaginary eigenvalues.
We now discuss three such networks that are inspired by different physical interpretations.

\subsection{The two-layer Hamiltonian network}

\cite{haber2017stable} propose a neural network architecture inspired by Hamiltonian systems 
\begin{equation}
\label{eq:one-layer-Hamiltonian}
  \begin{aligned}
  \dot{\mathbf{Y}}(t) &= \sigma(\mathbf{K}(t) \mathbf{Z}(t) + \mathbf{b}(t)), \\ 
  \dot{\mathbf{Z}}(t) &= -\sigma(\mathbf{K}(t)^T \mathbf{Y}(t) + \mathbf{b}(t)),
  \end{aligned}
\end{equation}
where $\mathbf{Y}(t)$ and $\mathbf{Z}(t)$ are partitions of the features, $\sigma$ is an activation function, and the network parameters are $\theta = (\mathbf{K},\mathbf{b})$. For convolutional neural networks, $\mathbf{K}(t)$ and $\mathbf{K}(t)^T$ are convolution operator and convolution transpose operator respectively.
It can be shown that the Jacobian matrix of this ODE satisfies the condition in Eq.~\eqref{equ:RJacobian}, thus it is stable and well-posed. The authors also demonstrate the performance on a small dataset.
However, in our numerical experiments we have found that the representability of this ``one-layer'' architecture is limited.

According to the universal approximation theorem \cite{hornik1991approximation}, a two-layer neural network can approximate any monotonically-increasing continuous function on a compact set. Recent work \cite{zhang2016understanding} shows that simple two-layer neural networks already have perfect finite sample expressivity as soon as the number of parameters exceeds the number of data points. 
Therefore, we propose to extend Eq.~\eqref{eq:one-layer-Hamiltonian} to the following two-layer structure:
\begin{equation}
\label{equ:two-layer-hamiltonian-simple}
\begin{aligned}
\dot{\mathbf{Y}}(t) &= \mathbf{K}_1^T(t) \sigma(\mathbf{K}_1(t) \mathbf{Z}(t) + \mathbf{b}_1(t)), \\
\dot{\mathbf{Z}}(t) &= -\mathbf{K}_2^T(t) \sigma(\mathbf{K}_2(t) \mathbf{Y}(t) + \mathbf{b}_2(t)).
\end{aligned}
\end{equation}
In principle, any linear operator can be used within the Hamiltonian framework. However, since our numerical experiments consider 
images, we choose
 $\mathbf{K}_i$ to be a convolution operator, $\mathbf{K}_i^T$ as its transpose.
Rewriting Eq.~\eqref{equ:two-layer-hamiltonian-simple} in matrix form gives
\begin{equation}
\label{eqn:two-layer-Hamiltonian-matrix-form}
\begin{pmatrix}
\dot{\mathbf{Y}} \\ 
\dot{\mathbf{Z}}
\end{pmatrix}
=
\begin{pmatrix}
\mathbf{K}_1^T & 0 \\ 
0 & -\mathbf{K}_2^T
\end{pmatrix}
\sigma
\Big(
\begin{pmatrix}
0 & \mathbf{K}_1 \\ 
\mathbf{K}_2 & 0
\end{pmatrix}
\begin{pmatrix}
\mathbf{Y} \\ 
\mathbf{Z}
\end{pmatrix}
+
\begin{pmatrix}
\mathbf{b}_1 \\ 
\mathbf{b}_2
\end{pmatrix}
\Big).
\end{equation}
There are different ways of partitioning the input features, including checkerboard partition and channel-wise partition \cite{dinh2016density}. In this work, we use equal channel-wise partition, that is, the first half of the channels of the input is $\mathbf{Y}$ and the second half is $\mathbf{Z}$.

It can be shown that the Jacobian matrix of Eq.~\eqref{eqn:two-layer-Hamiltonian-matrix-form} satisfies the condition in Eq.~\eqref{equ:RJacobian}, that is,
\begin{align}
\label{eqn:Jacobian-sandwich}
\mathbf{J} &= 
\nabla_{
\binom{\mathbf{Y}}{\mathbf{Z}}
}
\begin{pmatrix}
\mathbf{K}_1^T & 0 \\ 
0 & -\mathbf{K}_2^T
\end{pmatrix}
\sigma
\Big(
\begin{pmatrix}
0 & \mathbf{K}_1 \\ 
\mathbf{K}_2 & 0
\end{pmatrix}
\begin{pmatrix}
\mathbf{y} \\ 
\mathbf{z}
\end{pmatrix}
\Big)
\nonumber\\
&= \begin{pmatrix}
\mathbf{K}_1^T & 0 \\ 
0 & -\mathbf{K}_2^T
\end{pmatrix}
\mathrm{diag}(\sigma')
\begin{pmatrix}
0 & \mathbf{K}_1 \\ 
\mathbf{K}_2 & 0
\end{pmatrix},
\end{align}
where ${\rm diag}(\sigma')$ is the derivative of the activation function. 
The eigenvalues of $\mathbf{J}$ are all imaginary (see the Appendix for a proof).
Therefore Eq.~\eqref{equ:RJacobian} is satisfied and the forward propagation of our neural network is stable and well-posed. 

\begin{figure*}
	\begin{center}
			\includegraphics[width = 1\linewidth]{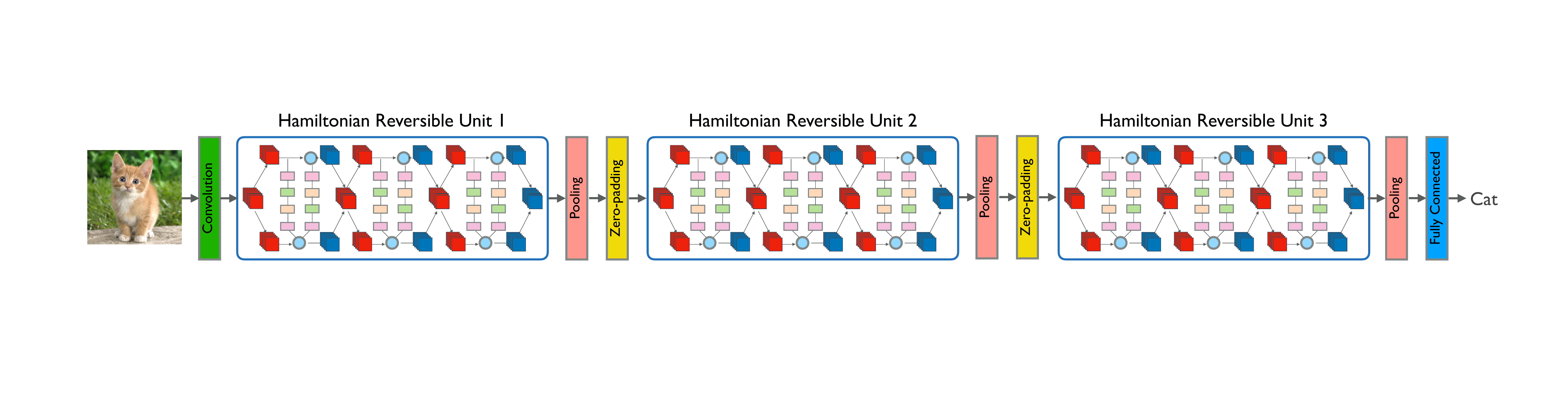}  
	\end{center}
	\caption{\textbf{The Hamiltonian Reversible Neural Network.} It is the simple stacking of several Hamiltonian reversible blocks as shown in Fig. \ref{fig:hamiltonian-reversible-block} and pooling layer. }
	\label{fig:hamiltonian-full}
\end{figure*}

A commonly used discretization technique for Hamiltonian systems such as Eq.~\eqref{equ:two-layer-hamiltonian-simple} is the Verlet method~\cite{ap} that reads
\begin{equation}
\label{eqn:hamiltonian-discretized}
\begin{aligned}
\mathbf{Y}_{j+1} &= \mathbf{Y}_{j} + h \mathbf{K}_{j1}^T \sigma(\mathbf{K}_{j1} \mathbf{Z}_{j} + \mathbf{b}_{j1}), \\
\mathbf{Z}_{j+1} &= \mathbf{Z}_{j} - h \mathbf{K}_{j2}^T \sigma(\mathbf{K}_{j2} \mathbf{Y}_{j+1} + \mathbf{b}_{j2}).
\end{aligned}
\end{equation}
We choose Eq.~\eqref{eqn:hamiltonian-discretized} to be our Hamiltonian blocks and illustrate it in Fig.~\ref{fig:hamiltonian-reversible-block}. 
Similar to ResNet \cite{he2016deep}, our Hamiltonian reversible network is built by first concatenating blocks to units, and then concatenating units to a network.
An illustration of our architecture is provided in Fig.~\ref{fig:hamiltonian-full}.

\subsection{The midpoint network}

Another reversible numerical method for discretizing the first-order ODE in Eq.~\eqref{equ:ODE} is obtained by using central finite differences in time
\begin{equation}
\frac{\mathbf{Y}_{j+1}-\mathbf{Y}_{j-1}}{2 h} = \mathcal{F}(\mathbf{Y}_j).
\end{equation}
This gives the following forward propagation
\begin{equation}
% \label{eqn:mid-point-discretization}
\mathbf{Y}_{j+1} = \mathbf{Y}_{j-1} + 2h\mathcal{F}(\mathbf{Y}_j),\; \text{ for }\; j=1,\ldots, N-1, 
\end{equation}
where $\mathbf{Y}_1$ is obtained by one forward Euler step. To guarantee stability for a single layer we can use the function ${\cal F}$ to contain an anti-symmetric linear operator, that is,
\begin{equation}
 \mathcal{F}(\mathbf{Y}) = \sigma((\mathbf{K} - \mathbf{K}^T) \mathbf{Y} + \mathbf{b}). 
\end{equation}
The Jacobian of this forward propagation is 
\begin{equation}
{\bf J} =  {\rm diag}(\sigma')(\mathbf{K} - \mathbf{K}^T),
\end{equation}
which has only imaginary eigenvalues.
This yields the single layer midpoint network
\begin{equation}
\label{eqn:mid-point-two-layer}
  \bfY_{j+1}=\left\{ \begin{aligned}
   &2h  \sigma((\mathbf{K}_{j} - \mathbf{K}_{j}^T) \mathbf{Y}_{j} + \mathbf{b}_{j}), & j=0,\\
   &\mathbf{Y}_{j-1} + 2h  \sigma((\mathbf{K}_{j} - \mathbf{K}_{j}^T) \mathbf{Y}_{j} + \mathbf{b}_{j}), &{\rm j>0.}
  \end{aligned}\right.
\end{equation}

As we see next, it is straightforward to show that the midpoint method is reversible (at least algebraically).
However,
while it is possible to potentially use a double layer midpoint network, it is difficult to
ensure the stability of such network. To this end, we explore the leapfrog network next.

\subsection{The leapfrog network}

A stable leapfrog network can be seen as a special case of the Hamiltonian network in Eq.~\eqref{equ:two-layer-hamiltonian-simple}  when one of the kernels is the identity matrix and one of the activation is the identity function. The leapfrog network involves two derivatives in time and reads 
\begin{equation}
\label{equ:second-ODE}
\ddot{\mathbf{Y}}(t)=  -\bfK(t)^T\sigma (\mathbf{K}(t)\bfY(t)+b(t)),\quad \bfY(0)=\bfY_0.
\end{equation}
It can be discretized, for example, using the conservative leapfrog discretization, which uses the following symmetric approximation to the second derivative in time
$$ \ddot{\bfY}(t_j) \approx h^{-2}( \bfY_{j+1} - 2 \bfY_j + \bfY_{j-1}). $$
Substituting the approximation in Eq.~\eqref{equ:second-ODE}, we obtain:
\begin{equation}
\label{eq:leapfrog-discretization}
  \bfY_{j+1}=\left\{ \begin{aligned}
   &2\bfY_j-h^2\bfK^T_j\sigma(\mathbf{K}_j \bfY_j+\mathbf{b}_j), & j=0,\\
   &2\bfY_j-\bfY_{j-1}-h^2\bfK_j^T\sigma(\mathbf{K}_j\bfY_j+\mathbf{b}_j), &{j>0.}
  \end{aligned}\right.
\end{equation}

\subsection{Reversible architectures and stability}

An architecture is called reversible if it  allows the reconstruction of the
activations going from the end to the beginning. Reversible numerical methods for ODEs have been studied
in the context of hyperbolic differential equations \cite{nguyen2014five}, and reversibility was discovered recently in the machine learning
community \cite{dinh2014nice,gomez2017reversible}.
Reversible techniques enable memory-efficient implementations of the network that requires the storage of the last activations only.

Let us first demonstrate the reversibility of the leapfrog network.
Assume that we are given the last two states, $\bfY_N$ and $\bfY_{N-1}$.
Then, using Eq.~\eqref{eq:leapfrog-discretization} it is straight-forward to compute $\bfY_{N-2}$:
\begin{multline}
\bfY_{N-2} = 2\bfY_{N-1} - \bfY_N \\ 
- h^2 \bfK_{N-1}^T \sigma(\bfK_{N-1}\bfY_{N-1} +\mathbf{b}_{N-1}). 
\end{multline}
Given $\bfY_{N-2}$ one can continue and re-compute the activations at each hidden layer during backpropagation. 
Similarly, it is straightforward to show that the midpoint network is reversible.

The Hamiltonian network is similar to the 
RevNet and can be described as
\begin{equation}
\label{eq:revnet-forward}
  \begin{aligned}
  \mathbf{Y}_{j+1} &=  \mathbf{Y}_{j} + \mathcal{F}(\mathbf{Z}_{j}), \\ 
  \mathbf{Z}_{j+1} &=  \mathbf{Z}_{j} + \mathcal{G}(\mathbf{Y}_{j+1}),
  \end{aligned}
\end{equation}
where $\mathbf{Y}_{j}$ and $\mathbf{Z}_{j}$ are a partition of the units in block $j$; $\mathcal{F}$ and $\mathcal{G}$ are the residual functions. 
% The transformation described by Eq.~\eqref{eq:revnet-forward} has non-unit determinant and therefore could represent any mappings rather than just volume-preserving mappings as in \cite{dinh2014nice}.  
Eq.~\eqref{eq:revnet-forward} is reversible as each layer's activations can be computed from the next layer's  as follows:
\begin{equation}
\label{eq:revnet-backward}
  \begin{aligned}
  \mathbf{Z}_{j} &=  \mathbf{Z}_{j+1} - \mathcal{G}(\mathbf{Y}_{j+1}), \\
  \mathbf{Y}_{j} &=  \mathbf{Y}_{j+1} - \mathcal{F}(\mathbf{Z}_{j}). \\ 
  \end{aligned}
\end{equation}
It is clear that Eq.~\eqref{eqn:hamiltonian-discretized} is a special case of Eq.~\eqref{eq:revnet-forward}, which enables us to implement Hamiltonian network in a memory efficient way.

While RevNet and MidPoint represent reversible networks algebraically, they may not be reversible in practice without restrictions on the residual functions.
To illustrate, consider the simple linear case where
${\cal G}(\bfY) = \alpha \bfY$ and ${\cal F}(\bfZ) = \beta \bfZ$.
% $$ {\cal G}(\bfY) = \alpha \bfY \quad {\rm and} \quad {\cal F}(\bfZ) = \beta \bfZ. $$
The RevNet in this simple case reads
\begin{equation*}
  \begin{aligned}
  \mathbf{Y}_{j+1} &=  \mathbf{Y}_{j} + \beta\mathbf{Z}_{j}, \\ 
  \mathbf{Z}_{j+1} &=  \mathbf{Z}_{j} + \alpha\mathbf{Y}_{j+1}. \\
  \end{aligned}
\end{equation*}
One way to simplify the equations is to look at two time steps and subtract them:
% \begin{equation*}
%   \begin{aligned}
%   \mathbf{Y}_{j+1} - \mathbf{Y}_{j}  &=   \beta\mathbf{Z}_{j} \\
%   \mathbf{Y}_{j} - \mathbf{Y}_{j-1}  &=   \beta\mathbf{Z}_{j-1}.
%   \end{aligned}
% \end{equation*}
% Subtracting both equations we obtain
$$ \mathbf{Y}_{j+1} - 2 \mathbf{Y}_{j} + \mathbf{Y}_{j-1} = \beta ({\bfZ}_{j} - {\bfZ}_{j-1}) = \alpha \beta \bfY_{j}, $$
which implies that
$$ \mathbf{Y}_{j+1} - (2 + \alpha \beta) \mathbf{Y}_{j} +  \bfY_{j-1} = 0.$$
These type of equations have a solution of the
form $\bfY_j = \xi^j$. The characteristic equation is
\begin{equation}\label{eq:xi}
	\xi^{2} - (2 + \alpha \beta) \xi  + 1  = 0.
\end{equation}
Define $a = \frac 12 (2 + \alpha \beta)$, the roots of the equation are
$ \xi = a \pm \sqrt{a^2-1}. $
If $a^2 \le 1$ then we have that
$\xi = a \pm i \sqrt{1-a^2}.$
and
$ |\xi|^2 =1, $
which implies that the method is stable and no energy in the feature vectors is added or lost.

It is obvious that Eq.~\eqref{eq:xi} is not stable for every choice of $\alpha$ and $\beta$.
Indeed, if, for example, $\alpha$ and $\beta$ are positive then  $|\xi| > 1$  and the solution
can grow at every layer exhibiting unstable behavior. It is possible
to obtain stable solutions if $0<\alpha$ and  $\beta<0$ and both are sufficiently small. This is the role of $h$
in our Hamiltonian network.

This analysis plays a key role in reversibility. For unstable networks, either the forward or the backward propagation consists of an exponentially growing mode. For computation in single precision (like most practical CNN), the gradient can be grossly inaccurate. Thus we see that not every choice of the functions ${\cal F}$ and ${\cal G}$ lead to a reasonable network in practice and that some control is needed if we are to have a network that does not
grow exponentially neither forward nor backwards.

\subsection{Arbitrarily deep residual neural networks}
All three architectures we proposed can be used with arbitrary depth, since they do not have any dissipation. This implies that the signal that is input into the system does not decay even for arbitrarily long networks. Thus signals can propagate through this system  to infinite network depth. We have also experimented with slightly dissipative networks, that is, networks that attenuate the signal at each layer, that yielded results that were comparable to the ones obtained by the networks proposed here.

% \subsection{Batch normalization}
% Batch normalization is a technique to accelerate the training of deep network by stabilizing the distribution of each layer's inputs \cite{ioffe2015batch}. It makes the networks more robust to bad initialization and it has become a common practice to use batch normalization in neural networks. However, several researchers have recently proposed methods without using batch normalization. For example, the critic network of WGAN-GP \cite{gulrajani2017improved} does not use batch normalization as the critic maps a single input to a single output while mapping from an entire batch of inputs to a batch of outputs using batch normalization makes the training objective no longer valid.
% In \cite{ye2017importance}, the authors also suggest that batch normalization is not appropriate for second-order optimization methods.

% If we regard a forward neural network as an ODE, batch normalization changes the dynamics of the differential equation: in each step, instead of mapping an input to an output, a batch of inputs is mapped to a batch of outputs. As a result, the Jacobian matrix no longer has a simple form. Since our Hamiltonian network is stable, there is no need to adjust the distribution of each layer's input. Therefore we omit batch normalization to avoid the overhead, enabling a faster training procedure.

\subsection{Regularization}
Regularization plays a vital role serving as parameter tuning in the deep neural network training to help improve generalization performance \cite{zhang2016understanding}. Besides commonly used weight decay, we also use weight smoothness decay. Since we interpret the forward propagation of our Hamiltonian network as a time-dependent nonlinear dynamic process, we prefer convolution weights $\mathbf{K}$ that are smooth in time by using the regularization functional
\begin{equation*}
R(\mathbf{K})= \int_0^T  \|\dot{\mathbf{K}}_1(t)\|_F^2 + \|\dot{\mathbf{K}}_2(t)\|_F^2 \ \mathrm{d}t,
\end{equation*}
where $\|\cdot\|_F$ represents the Frobenius norm. 
Upon discretization, this gives the following weight smoothness decay as a regularization function
\begin{equation}
R_h(\mathbf{K})= h \sum_{j=1}^{T-1} \sum_{k=1}^{2} \left\|\frac{\mathbf{K}_{jk}-\mathbf{K}_{j+1,k}}{h}\right\|_F^2.
\end{equation}

\section{Experiments}\label{sec:exp}

We evaluate our methods on three standard classification benchmarks  (CIFAR-10, CIFAR100 and STL10) and compare against state-of-the-art results from the literature. Furthermore, we investigate the robustness of our method as the amount of training data decrease and train a deep network with 1,202 layers.

\begin{table*}
\begin{center}
\begin{tabular}{|l|l|l|c|c|c|c|}
\hline
\textbf{Name}   & \textbf{Units} & \textbf{Channels} & \multicolumn{2}{|c|}{\textbf{\# Model Params (M)}} &\multicolumn{2}{|c}{\textbf{Accuracy}} \vline\\
\hline
& & & \textbf{CIFAR-10}  & \textbf{CIFAR-100}& \textbf{CIFAR-10} &\textbf{CIFAR-100} \\
\hline 
\hline 
\textbf{ResNet-32} &5-5-5 &16-32-64 & 0.46& 0.47 & 92.86\% &70.05\% \\
\textbf{RevNet-38} &3-3-3 &32-64-112 &0.46& 0.48 & 92.76\%& 71.04\%  \\
%\textbf{HrNet-38 (Ours)} & 3-3-3 & 32-64-112 & 0.19 & 0.20 & 8.46\% & 32.88\% \\
\textbf{Hamiltonian-74 (Ours)} & 6-6-6 & 32-64-112 & 0.43 & 0.44 & 92.76\% & 69.78\% \\
% \textbf{MidPoint-32 (Ours)} & 5-5-5 & 16,32,64 & 0.20 & - & - \%& - \% \\
% \textbf{MidPoint-20w (Ours)} & 3-3-3 & 32-64-112 & 0.34 & - & 9.12\%& - \% \\
\textbf{MidPoint-26 (Ours)} & 4-4-4 & 32-64-112 & 0.50 & 0.51 & 91.16\%& 67.25\% \\

% \textbf{MidPoint-62 (Ours)} & 10-10-10 & 16,32,64 & 0.45 & - & 10.65\% & - \% \\
% \textbf{SecondOrder-62 (Ours)} & 10-10-10 & 16,32,64 & 0.45 &  & 12.54\% & - \% \\
\textbf{Leapfrog-26 (Ours)} & 4-4-4 & 32-64-112 & 0.50 & 0.51 & 91.92\% & 69.14\% \\
% \textbf{HrNet-50 (Ours)} & 3-3-3-3 &16-32-64-80 &0.46 &0.46&8.45\% &32.18\% \\
% \textbf{MidPoint-50 (Ours)} &3-3-3-3&16-32-64-80 &0.44 &0.45 &8.89\% &33.53\%\\
%\textbf{SecondOrder-50 (Ours)} & 3-3-3-3 & 16-32-64-80 &0.33& 0.34& 9.92\% &34.52\%\\ 
% \textbf{SecondOrder-26 (Ours)} & 3-3-3-3 & 16-32-64-112 &0.50&0.51 & 90.83\% &66.69\%\\ 
% \textbf{SecondOrder-34 (Ours)} & 4-4-4-4 & 16-32-64-112 &0.51 & - & Lili running\% & Lili running\%\\ 
\hline
\textbf{ResNet-110} & 18-18-18 & 16-32-64& 1.73 &1.73 & 94.26\%&73.56\%\\
\textbf{RevNet-110}  &9-9-9& 32-64-128& 1.73& 1.74 & 94.24\%&74.60\%\\
% \textbf{HrNet-110 (Ours)} & 9-9-9 &32-64-128 &0.81 &0.82 &6.80\%&29.57\%\\
\textbf{Hamiltonian-218 (Ours)} & 18-18-18 &32-64-128 &1.68 &1.69 & 94.02\%& 73.89\%\\
\textbf{MidPoint-62 (Ours)} & 10-10-10 & 32-64-128 & 1.78 & 1.79 & 92.76\% & 70.98\% \\
\textbf{Leapfrog-62 (Ours)} & 10-10-10 & 32-64-128 & 1.78 & 1.79 & 93.40\% & 72.28\% \\
% \textbf{MidPoint-218 (Ours)} &36-36-36 & 16,32,64 & & - & -\% & - \% \\
% \textbf{SecondOrder-62w (Ours)} & 10-10-10 & 32-64-128 & 1.78 & 1.79 & 91.85\% & 67.35\% \\
% \textbf{SecondOrder-62 (Ours)} & 12-12-12-12 & 16-32-64-112 & 1.80 & - & Bo running\% & -\% \\
% \textbf{SecondOrder-62w (Ours)} & 10-10-10 & 32-64-128 & 1.78 & 1.79 & 91.85\% & 67.35\% \\
% \textbf{SecondOrder-218 (Ours)} & 36-36-36 & 16,32,64 & 1.71 & - & 11.47\% & - \% \\
\hline
\textbf{ResNet-1202} &200-200-200 & 32-64-128 &19.4 & - & 92.07\%& -  \\
\textbf{Hamiltonian-1202 (Ours)} &100-100-100 & 32-64-128 &9.70 & - & 93.84\%& -  \\
\hline
\end{tabular}
\caption{\textbf{Main results for different architectures on CIFAR-10 and CIFAR-100.} We compare our three dynamical system inspired neural networks (Hamiltonian, MidPoint, and Leapfrog) with the state-of-the-art methods ResNet \cite{he2016deep} and RevNet \cite{gomez2017reversible}. Please note RevNet and our three architectures are much more memory-efficient than ResNet.
}
\label{tab:Hr-architecture}
\end{center}
\end{table*}

\subsection{Datasets and baselines}

\subsubsection{CIFAR-10 and CIFAR-100 }
The CIFAR-10 dataset \cite{krizhevsky2009learning} consists of 50,000 training images and 10,000 testing images in 10 classes with $32\times32$ image resolution.  The CIFAR-100 dataset uses the same image data and train-test split as CIFAR-10, but has 100 classes. 
We use the common data augmentation techniques including padding four zeros around the image, random cropping, random horizontal flipping and image standardization. Two state-of-the-art methods ResNet \cite{he2016deep} and RevNet \cite{gomez2017reversible} are used as our baseline methods.

\subsubsection{STL-10}
The STL-10 dataset \cite{coates2011analysis} is an image recognition dataset with 10 classes at image resolutions of $96\times 96$. It contains 5,000 training images and 8,000 test images. Thus, compared with CIFAR-10, each class has fewer labeled training samples but higher image resolution.
We used the same data augmentation as the CIFAR-10 except padding $12$ zeros around the images.

We use three state-of-the-art methods as baselines for the STL-10 dataset: Deep Representation Learning \cite{yang2015deep}, Convolutional Clustering \cite{dundar2015convolutional}, and Stacked what-where auto-encoders \cite{journals/corr/ZhaoMGL15}. 

% \subsubsection{Street View House Numbers (SVHN):}  SVHN \cite{netzer2011reading} is a real-world image dataset for developing machine learning and object recognition algorithms with minimal requirement on data preprocessing and formatting. It is obtained from house numbers in Google Street View images, and contains over 600,000 digit images with $32\times32$ image resolution. Following the common practice, data augmentation is not used for this dataset.

% We use four state-of-the-art methods as baselines for SVHN datasets: RNN for object recognition \cite{liang2015recurrent}, Generalizing Pooling Functions in CNN \cite{lee2016generalizing}, ResNet \cite{he2016deep}, ResNet with Stochastic Depth \cite{huang2016deep}.

\subsection{Neural network architecture specifications}
We provide the neural network architecture specifications here. The implementation details are in the Appendix.
All the networks contain 3 units, and each unit has $n$ blocks. There is also a convolution layer at the beginning of the network and a fully connected layer in the end. For Hamiltonian networks, there are 4 convolution layers in each block, so the total number of  layers is $12n+2$. For MidPoint and Leapfrog, there are 2 convolution layers in each block, so the total number of layers is $6n+2$.
In the first block of each unit, the feature map size is halved and the number of filters is doubled. We perform downsampling by average pooling and increase the number of filters by padding zeros. 

\subsection{Main Results and Analysis}

\begin{table}
\begin{center}
\begin{tabular}{|l|l|c|}
\hline
&\textbf{Methods}  & \textbf{Accuracy}  \\
\hline 
\hline
\textbf{Baselines}&\textbf{\cite{yang2015deep}} &73.15\%  \\
&\textbf{(Dundar et al. 2015)} & 74.1\% \\
&\textbf{\cite{journals/corr/ZhaoMGL15}} & 74.3\% \\ 
\hline
\hline
\textbf{Ours} &
\textbf{Hamiltonian} & 85.5\% \\
&\textbf{MidPoint}  & 84.6\% \\
&\textbf{Leapfrog} & 83.7\% \\
\hline
\end{tabular}
\caption{\textbf{Main results on STL-10.} All our three architectures outperform the benchmark methods by about 10\%.}
\label{tab:STL-10}
\end{center}
\end{table}

\subsubsection{CIFAR-10 and CIFAR-100} We show the main results of different architectures on CIFAR-10/100 in Table \ref{tab:Hr-architecture}. Our three architectures achieve comparable performance with ResNet and RevNet in term of accuracy using similar number of model parameters. Compared with ResNet, our architectures are more memory efficient as they are reversible, thus we do not need to store activations for most layers. While compared with RevNet, our models are not only reversible, but also stable, which is theoretically proved in Sec. \ref{sec:meth}. We  show later that the stable property makes our models more robust to small amount of training data and arbitrarily deep.

\subsubsection{STL-10} Main results on STL-10 are shown in Table \ref{tab:STL-10}. Compared with the state-of-the-art results, all our architectures achieve better accuracy.

\subsection{Robustness to training data subsampling}
Sometimes labeled data are very expensive to obtain. Thus, it is desirable to design architectures that generalize well when trained with few examples. To verify our intuition that stable architectures generalize well, we conducted extensive numerical experiments using the CIFAR-10 and STL-10 datasets with decreasing number of training data. Our focus is on the behavior of our neural network architecture in face of this data subsampling, instead of improving the state-of-the-art results. Therefore we intentionally use simple architectures:
4 blocks, each has 4 units, and the number of filters are $16-64-128-256$.
For comparison, we use ResNet~\cite{he2016deep} as our baseline.  CIFAR-10 has much more training data than STL-10 (50,000 vs 5,000), so we randomly subsample the training data from $20\%$ to $0.05\%$ for CIFAR-10, and from $80\%$ to $5\%$ for STL-10. The test data set remains unchanged.

\begin{figure}
	\begin{center}
		\includegraphics[width = 0.8\linewidth]{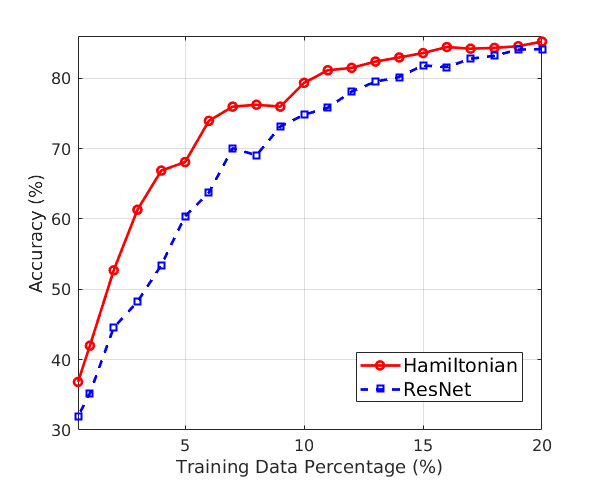}  
	\end{center}
	\caption{\textbf{Hamiltonian vs ResNet test accuracy on CIFAR10 with a small subset of training data.} }
	\label{fig:HrNet-CIFAR10-smallData}
\end{figure}

\subsubsection{CIFAR-10} Fig.~\ref{fig:HrNet-CIFAR10-smallData} shows the result on CIFAR-10 when decreasing the number examples in the training data from $20\%$ to $5\%$. Our Hamiltonian network performs consistently better in terms of accuracy than ResNet, achieving up to $13\%$ higher accuracy  when trained using just $3\%$ and $4\%$ of the original training data.

\subsubsection{STL-10} 
\begin{figure}
	\begin{center}
		\includegraphics[width = 0.8\linewidth]{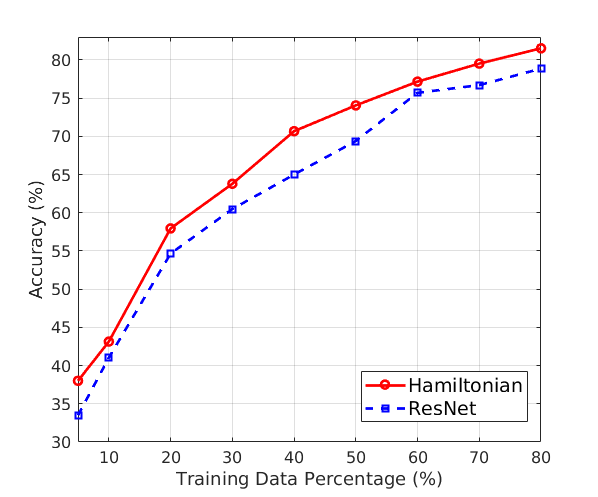}  
	\end{center}
	\caption{\textbf{Hamiltonian vs ResNet test accuracy for STL10 with a small subset of training data.} }
	\label{fig:HrNet-STL10}
\end{figure}

From the result as shown in Fig.~\ref{fig:HrNet-STL10}, we see that Hamiltonian consistently achieves better accuracy than ResNet with the average improvement being around $3.4\%$. Especially when using just $40\%$ of the training data, Hamiltonian has a $5.7\%$ higher accuracy compared to ResNet.

\subsection{Training a 1202-layer Hamiltonian}
To demonstrate the stability and memory-efficiency of the Hamiltonian network with arbitrary depth, we explore a 1202-layer architecture on CIFAR-10.
An aggressively deep ResNet is also trained on CIFAR-10 in \cite{he2016deep} with 1202 layers, which has an accuracy of $92.07\%$. 
Our result is shown at the last row of Table~\ref{tab:Hr-architecture}. Compared with the original ResNet, our architecture uses only a half of parameters and obtains better accuracy. 
Since the Hamiltonian network is intrinsically stable, it is guaranteed that there is no issue of exploding or vanishing gradient.
We can easily train an arbitrarily deep Hamiltonian network without any difficulty of optimization. The implementation of our reversible architecture is memory efficient, which enables a 1202 layer Hamiltonian model running on a single GPU machine with 10GB GPU memory.  

\section{Conclusion}\label{sec:conc}
We present three stable and reversible architectures that connect the stable ODE with deep residual neural networks and yield well-posed learning problems. We exploit the intrinsic reversibility property to obtain a memory-efficient implementation, which does not need to store the activations at most of the hidden layers. Together with the stability of the forward propagation, this allows training deeper architectures with limited computational resources. We evaluate our methods on three publicly available datasets against several state-of-the-art methods. Our experimental results demonstrate the efficacy of our method with superior or on-par state-of-the-art performance. Moreover, with small amount of training data, our architectures achieve better accuracy compared with the widely used state-of-the-art ResNet. We attribute the robustness to small amount of training data to the intrinsic stability of our Hamiltonian neural network architecture.

\section{Appendix}

\subsection{Proof: All eigenvalues of $\mathbf{J}$ in Eq.~\eqref{eqn:Jacobian-sandwich} are imaginary.}
The Jacobian matrix $\mathbf{J}$ is defined in Eq.~\eqref{eqn:Jacobian-sandwich}.
% \begin{equation}
% \mathbf{J}
% = \begin{pmatrix}
% \mathbf{K}_1^T & 0 \\ 
% 0 & -\mathbf{K}_2^T
% \end{pmatrix}
% \mathrm{diag}(\sigma')
% \begin{pmatrix}
% 0 & \mathbf{K}_1 \\ 
% \mathbf{K}_2 & 0
% \end{pmatrix}.
% \end{equation}
If $\mathbf{A}$ and $\mathbf{B}$ are two invertible matrices of the same size, then $\mathbf{AB}$ and $\mathbf{BA}$ have the same eigenvalues (Theorem 1.3.22 in \cite{horn2012matrix}).
If we define
\begin{align}
\mathbf{J'} &= 
\mathrm{diag}(\sigma')
\begin{pmatrix}
0 & \mathbf{K}_1 \\ 
\mathbf{K}_2 & 0
\end{pmatrix}
\begin{pmatrix}
\mathbf{K}_1^T & 0 \\ 
0 & -\mathbf{K}_2^T
\end{pmatrix}
\nonumber\\
&=
\mathrm{diag}(\sigma')
\begin{pmatrix}
0 & -\mathbf{K}_1 \mathbf{K}_2^T \\ 
\mathbf{K}_2 \mathbf{K}_1^T & 0
\end{pmatrix}
\nonumber\\
&:=\mathbf{DM},
\end{align}
then $\mathbf{J}$ and $\mathbf{J}'$ have the same eigenvalues. 
% Denote $\mathbf{D} = \mathrm{diag}(\sigma')$ and 
% $$
% \mathbf{M} = \begin{pmatrix}
% 0 & -\mathbf{K}_1 \mathbf{K}_2^T \\ 
% \mathbf{K}_2 \mathbf{K}_1^T & 0
% \end{pmatrix},
% $$
% then 
$\mathbf{D}$ is a diagonal matrix with non-negative elements, and $\mathbf{M}$ is a real anti-symmetric matrix such that $\mathbf{M}^T = -\mathbf{M}$.
Let $\lambda$ and $v$ be a pair of eigenvalue and eigenvector of $\mathbf{J'}=\mathbf{D}\mathbf{M}$, then
\begin{align}
\mathbf{D}\mathbf{M} \mathbf{v} &= \lambda \mathbf{v}, \\
\mathbf{M} \mathbf{v} &= \lambda \mathbf{D}^{-1} \mathbf{v}, \\
\mathbf{v}^{*} \mathbf{M} \mathbf{v} &= \lambda (\mathbf{v}^{*} \mathbf{D}^{-1} \mathbf{v}),
\end{align}
where $\mathbf{D}^{-1}$ is the generalized inverse of $\mathbf{D}$.
On one hand, since $\mathbf{D}^{-1}$ is non-negative definite, $\mathbf{v}^{*} \mathbf{D}^{-1} \mathbf{v}$ is real.
On the other hand, 
\begin{equation}
\label{eqn:vMv-imaginary}
(\mathbf{v}^* \mathbf{M} \mathbf{v})^* = \mathbf{v}^* \mathbf{M}^*\mathbf{v} = \mathbf{v}^* \mathbf{M}^T \mathbf{v} = -\mathbf{v}^* \mathbf{M}\mathbf{v},
\end{equation}
where $*$ represents conjugate transpose. Eq.~\ref{eqn:vMv-imaginary} implies that $\mathbf{v}^* \mathbf{M} \mathbf{v}$ is imaginary. Therefore, $\lambda$ has to be imaginary. As a result, all eigenvalues of $\mathbf{J}$ are imaginary.

\subsection{Implementation details}
Our method is implemented using TensorFlow library \cite{abadi2016tensorflow}. The CIFAR-10/100 and STL-10 experiments are evaluated on a desktop with an Intel Quad-Core i5 CPU and a single Nvidia 1080 Ti GPU.

For CIFAR-10 and CIFAR-100 experiments, we use a fixed mini-batch size of 100 both for training and test data except Hamiltonian-1202, which uses a batch-size of 32. The learning rate is initialized to be 0.1 and decayed by a factor of 10 at 80, 120 and 160 training epochs. The total training step is 80K. The weight decay constant is set to $2\times10^{-4}$, weight smoothness decay is $2\times10^{-4}$ and the momentum is set to 0.9.

For STL-10 experiments, the mini-batch size is 128. The learning rate is initialized to be 0.1 and decayed by a factor of 10 at 60, 80 and 100 training epochs. The total training step is 20K. The weight decay constant is set to $5\times10^{-4}$, weight smoothness decay is $3\times10^{-4}$ and the momentum is set to 0.9.

{\small
\bibliographystyle{aaai}
\bibliography{egbib}
}

\end{document}